\documentclass[conf]{new-aiaa}
\usepackage{titlesec}
\usepackage[utf8]{inputenc}
\usepackage{graphicx,color}
\usepackage{amsmath}
\usepackage[version=4]{mhchem}
\usepackage{siunitx}
\usepackage{longtable,tabularx}
\usepackage{subfigure}
\expandafter\def\csname ver@subfig.sty\endcsname{}
\usepackage{float}
\usepackage{subfig}
\usepackage{tikz}
\usepackage{forest}
\usepackage{stanli}

\newcommand{\tstar}[5]{
\pgfmathsetmacro{\starangle}{360/#3}
\draw[#5] (#4:#1)
\foreach \x in {1,...,#3}
{ -- (#4+\x*\starangle-\starangle/2:#2) -- (#4+\x*\starangle:#1)
}
-- cycle;
}

\title{SMT-EX: An Explainable Surrogate Modeling Toolbox for Mixed-Variables Design Exploration}

\author{Mohammad Daffa Robani}
\affil{AwanTunai, Urban Suites, 3 Hullet Road, Singapore, 229158}
\author{Paul Saves\footnote{Ph.D, email: paul.saves@onera.fr}}
\affil{DTIS, ONERA, Université de Toulouse, 31000 Toulouse, France}
\affil{Fédération ENAC ISAE-SUPAERO ONERA, Université de Toulouse, 31000 Toulouse, France}
\author{Lavi Rizki Zuhal\footnote{Professor, Faculty of Mechanical and Aerospace Engineering, lavirz@ae.itb.ac.id}, Pramudita Satria Palar\footnote{Assistant Professor, Faculty of Mechanical and Aerospace Engineering, pramsp@itb.ac.id (Corresponding author)}}
\affil{Bandung Institute of Technology, Bandung, Indonesia, 40132}
\author{Joseph Morlier\footnote{Professor, email: joseph.morlier@isae-supaero.fr}}
\affil{ICA, Université de Toulouse, ISAE-SUPAERO, MINES ALBI, UPS, INSA, CNRS, 3 rue Caroline Aigle, \\ Toulouse, 31400, France}

\begin{document}

\maketitle

\begin{abstract}
Surrogate models are of high interest for many engineering applications, serving as cheap-to-evaluate time-efficient approximations of black-box functions to help engineers and practitioners make decisions and understand complex systems. As such, the need for explainability methods is rising and many studies have been performed to facilitate knowledge discovery from surrogate models.
To respond to these enquiries, this paper introduces SMT-EX, an enhancement of the open-source Python Surrogate Modeling Toolbox (SMT) that integrates explainability techniques into a state-of-the-art surrogate modelling framework.
More precisely, SMT-EX includes three key explainability methods: Shapley Additive Explanations, Partial Dependence Plot, and Individual Conditional Expectations.
A peculiar explainability dependency of SMT has been developed for such purpose that can be easily activated once the surrogate model is built, offering a user-friendly and efficient tool for swift insight extraction. The effectiveness of SMT-EX is showcased through two test cases. The first case is a 10-variable wing weight problem with purely continuous variables and the second one is a 3-variable mixed-categorical cantilever beam bending problem. Relying on SMT-EX analyses for these problems, we demonstrate its versatility in addressing a diverse range of problem characteristics. SMT-Explainability is freely available on Github\footnote{\url{https://github.com/SMTorg/smt-explainability}}.
\end{abstract}



\section{Introduction}
Numerous engineering tasks, including optimization, sensitivity analysis, and uncertainty quantification, often require multi-query calls that involve repeated use of computationally intensive solvers, such as high-fidelity computational fluid dynamics~\cite{farhat1998load}. Surrogate models are commonly employed to efficiently address multi-query tasks~\cite{yondo2018review}. These models serve as cost-effective approximations of black-box functions, constructed using a limited number of data points. The data is usually obtained from computer simulations (such as computational fluid dynamics or finite element simulations) or physical experiments. Essentially, it consists of pairs of data points and their corresponding responses. Some popular types of surrogate models in engineering literature include Gaussian Process Regression (GPR)~\cite{krige1951statistical}, polynomial chaos expansion~\cite{zuhal2021dimensionality} and neural networks~\cite{lecun2015deep}. The importance of the surrogate model in the modern engineering landscape cannot be overstated since it provides means to significantly reduce the turn-around time for more efficient optimization~\cite{he2023review, deb2020surrogate}, design exploration, and sensitivity analysis~\cite{cheng2020surrogate}, among others. 

Several open-source surrogate modeling tools have been developed to assist engineers in implementing surrogate models. In particular, the Surrogate Modeling Toolbox (SMT)~\cite{bouhlel2019python, saves2024smt} is an open-source framework offering functions to efficiently construct surrogate models. An important key surrogate model implemented in SMT is the GPR, also known as the Kriging model. 
The latter is a probabilistic framework for approximating a black-box function that can predict both a mean value prediction and an uncertainty quantification and can therefore be used for Monte Carlo computations~\cite{eliavs2020periodic} and variance-based sensitivity analysis~\cite{menz2021variance}. The versatility of GPR has proven useful for various enhancements that improve its ability to address different types of problems. To handle various input types, SMT has extended GPR to accommodate mixed-integer and mixed-categorical variables~\cite{saves_mixed}. As a surrogate modeling tool, various experiments have shown that the GPR module of SMT has better predictive power than other similar tools, \textit{e.g.}, UQlab~\cite{marelli2014uqlab} and scikit-learn for a fixed number of points~\cite{faraci2022review}. Thus, SMT shows promise for various real-world applications and has the potential to expand its capabilities further. This paper focuses on the extension of SMT to expand its capabilities beyond as a data-driven prediction tool. Specifically, this work enhances SMT functionalities by adding a knowledge discovery tool, which will be explained shortly.

Engineers frequently aim to derive meaningful insights from constructed surrogate models. For example, understanding the influence of input variables on the output can facilitate the reduction of the problem's dimensionality. Additionally, engineers often seek to determine the complexity of the input-output relationship, such as the degree of nonlinearity. These tasks are more easily accomplished when the problem exhibits low dimensionality in terms of input variables but it is not a trivial task to gain such an understanding for problems with high-dimensional input~\cite{saves_mixed_hd}. High accuracy is often gained by sacrificing the interpretability of the model, which makes knowledge extraction even more difficult. Additional efforts are then necessary to be able to gain understanding. To that end, explainability serves as a powerful means to ``dissect'' the input-output relationship and provide knowledge to the user~\cite{roscher2020explainable, belle2021principles, linardatos2020explainable}. The term explainability itself originated from the machine learning community, which aims to improve the transparency of the model and eventually enhance our trust in the model. However, in this paper, we particularly focus on the role of explainability in knowledge extraction and scientific understanding. Several explainability methods have been proposed, including SHapley Additive Explanations (SHAP)~\cite{lundberg2017unified}, Accumulated Local Effects (ALE)~\cite{apley2020visualizing}, Partial Dependence Plots (PDP)~\cite{friedman2001greedy}, and Individual Conditional Expectations (ICE)~\cite{goldstein2015peeking}. Each method reveals understanding from various viewpoints, which is why it is essential to deploy various methods when dealing with a single problem. While explainability originated in the machine learning and statistics community, methods from other fields, like Global Sensitivity Analysis (GSA)~\cite{rohmer2011global} and active subspace~\cite{tripathy2016gaussian}, also strive to offer insights into the constructed model. In particular, variance-based GSA methods like Sobol' indices aim to rank the significance of input variables based on specific rules, using variance decomposition as their foundation~\cite{sobol2001global}. An in-depth litterature on the subject can be find here~\cite{razavi2021future}.
Conversely, active subspace enables the identification and visualization of low-dimensional latent spaces within the input-output relationship. In the field of engineering design, the authors have made several contributions, including the analytical calculation of SHAP for PCE~\cite{palar2023enhancing} and the development of a multi-objective design exploration method using SHAP~\cite{palar2024multi}. We have also successfully applied explainability to various engineering design problems, such as for distributed electric propulsion aircraft design~\cite{palar2024design}.

In this paper, we present an extension of the SMT toolbox, named SMT-EX, which is specifically designed for explainability analysis. Our primary aim is to provide users with an easy-to-use explainability tool that can be readily deployed once the surrogate model has been constructed with SMT. The SMT-EX features several explainability submodules, including SHAP, PDP, ICE, and Sobol' indices. Thanks to the original capabilities of the SMT, the developed explainability module is also equipped with the ability to provide explainability for both continuous and mixed-categorical variables. \textcolor{black}{Besides, we have implemented the module for uncertainty quantification using conformal prediction in SMT. The uncertainty quantification subroutine itself can be seen as a part of ``explainability" since it provides a means to make the model transparent. It is based on a simple split prediction from~\cite{da2024tutorial}.}

In this paper, we demonstrate the applicability of SMT-EX for two test problems with various characteristics, namely, a 10-variable wing weight problem with all continuous variables and a 3-variable mixed-categorical cantilever beam problem.

\section{SMT 2.0: Surrogate Modeling Toolbox}
Let us denote the vector of input variable $\boldsymbol{x}=[\boldsymbol{x}_{1},\boldsymbol{x}_{2},\ldots,\boldsymbol{x}_{m}]^{T}$, where $m$ is the size of input variables. The aim is to approximate a black-box function $f(\boldsymbol{x})$ with a surrogate model $\hat{f}(\boldsymbol{x})$, where
\begin{equation}
    f:\Omega \times \mathbb{F}^{l} \rightarrow \mathbb{R},
\end{equation}
$\Omega \subset \mathbb{R}^{m_{c}}$ is the continuous design set for $m_{c}$ continuous variables, and $\mathbb{F}^{l} = \{1,\ldots,L_{1}\} \times \{1,\ldots,L_{2}\} \times \ldots \{1,\ldots,L_{l}\}$ is the input space for $l$ categorical variables with $L_{1},\ldots,L_{l}$ represent their respective levels.

SMT 2.0 is a Python package offering a comprehensive suite of surrogate modeling techniques\footnote{The SMT explainaiblity module is part of the SMTorg organization: \url{https://github.com/SMTorg}}, sampling strategies, and benchmarking tools. SMT features an intuitive library of surrogate models and facilitates the incorporation of new methods with ease. It is the product of collaborative efforts between various institutions, including ONERA, NASA Glenn, ISAE-SUPAERO/ICA, University of Michigan, Polytechnique Montréal, and the University of California San Diego. Several unique modules and capabilities of SMT include gradient-enhanced modeling and unique surrogate models (\textit{e.g.}, Kriging with partial least squares~\cite{bouhlel2016improving} and energy-minimizing spline interpolation~\cite{hwang2018fast}). The newest SMT version is SMT 2.0, with additional modules focusing on hierarchical and mixed-variables Gaussian processes. 
This toolbox is open-source, and, as such, can be used for optimization and understanding with other software such as SBArchOpt~\cite{bussemaker2023sbarchopt} or SEGOMOE~\cite{bartoli_sego} but these applications were limited to simple Monte Carlo and Bayesian optimization and need to be extended for explainability, especially when dealing with architecture problems, \textit{e.g.}, as discussed in~\cite{palar2024design}.

In this paper, with an emphasis on explainability, we focus on the GPR implemented in SMT. The strength of GPs lies in their ability to handle small to medium-sized data sets, typical of engineering problems that involve computationally expensive simulations during data generation. Moreover, SMT surrogate models module accommodates various types of input variables, including its latest capability to address mixed-categorical problems, which is pertinent to one of the problems discussed in this paper. It is worth noting that the explainability module in SMT-EX is versatile and can be integrated with any type of surrogate model in SMT, not just limited to GPR.

SMT has been used in various engineering applications, and it will be beneficial for the users to enhance its capability with an additional tool that can help users understand their models, as explained in the next section. However, SMT has heavily focused on GPR models and derivative predictions for application in surrogate-based modelling or optimization. As such, for a given point, we can, not only predict the value at this point but put this prediction within a context and quantify uncertainty among others. More details are given hereinafter. 
 

\subsection{Gaussian process predictions}
GPR are a powerful tool in surrogate modelling, especially for their ability to quantify uncertainty in predictions. A Gaussian Process (GP) provides a probabilistic model that represents the output as a collection of random variables with a joint Gaussian distribution.  This approach enables the estimation of both mean predictions and variance for each input, effectively providing a natural mechanism for uncertainty quantification.

Given a set of observations \(\mathcal{D} = \{(\boldsymbol{x}_i, y_i)\}_{i=1}^n\) where \(y_i = f(\boldsymbol{x}_i) + \epsilon_i\) (with \(\epsilon_i\) being Gaussian noise), the GP prior for the function \(f\) is specified by a mean function \(m(\boldsymbol{x})\) and a covariance function \(k(\boldsymbol{x}, \boldsymbol{x}')\). For most practical applications, we often assume a zero mean function, simplifying our focus to the covariance structure. The prediction at a new point \(\boldsymbol{x}_*\) can be expressed as:
\begin{align}
    \hat{y}_* & = m(\boldsymbol{x}_*) + k(\boldsymbol{x}_*, \boldsymbol{X})^T K^{-1} (\boldsymbol{y} - m(\boldsymbol{X})), \\
    \sigma_*^2 & = k(\boldsymbol{x}_*, \boldsymbol{x}_*) - k(\boldsymbol{x}_*, \boldsymbol{X})^T K^{-1} k(\boldsymbol{x}_*, \boldsymbol{X}),
\end{align}
where \(K\) is the covariance matrix computed from the training data, and \(k(\boldsymbol{x}_*, \boldsymbol{X})\) represents the covariance between the new point and the training points.
In particular, in the mixed categorical setting, this matrix depends on some other matrices of correlation between the levels of every  categorical variables~\cite{saves_mixed}. This matrices are used internally by the GP and automatically learn and as such, we propose to display them as they reflect the learned correlations between the levels of the inputs categorical variables~\cite{saves_mixed_hd}.
In a more supervised context, by coupling with a clustering algorithm,  one can impose a structure on the learned correlation hyperparameters to infer more easily "group of levels" that behave the same~\cite{roustant2020group}.

The confidence interval can be derived from the predictive distribution of \(y_*\), which is Gaussian with mean \(\hat{y}_*\) and variance \(\sigma_*^2\). A \(100(1-\alpha)\%\) confidence interval for a new observation \(y_*\) can then be expressed as:
\begin{equation}
    \hat{y}_* \pm z_{\alpha/2} \sigma_*, 
\end{equation}
where \(z_{\alpha/2}\) is the critical value from the standard normal distribution corresponding to the desired confidence level. This interval provides a range of plausible values for the prediction, incorporating both the uncertainty in the model and the inherent noise in the observations. GPs and their ability to generate confidence intervals have found applications in various fields, including engineering optimization and machine learning. Notably, the integration of GPs with other methods, such as conformal prediction, enhances their reliability and applicability in decision-making processes \cite{williams2006gaussian}.

\subsection{Split conformal predictions}

Conformal prediction is a powerful framework for quantifying the uncertainty of predictions made by machine learning models. Unlike traditional confidence intervals, conformal prediction enables adaptive and valid uncertainty quantification based on the specific data at hand. Split conformal prediction, in particular, enhances this framework by leveraging a calibration dataset to derive confidence intervals~\cite{barber2023conformal}.

The general approach of split conformal involves splitting the available dataset \(\mathcal{D}\) into two parts: a training set \(\mathcal{D}_{\text{train}} = \{(\boldsymbol{x}_i, y_i)\}_{i=1}^{n_{\text{train}}}\) for model fitting and a calibration set \(\mathcal{D}_{\text{cal}} = \{(\boldsymbol{x}_j, y_j)\}_{j=1}^{n_{\text{cal}}}\) for assessing the model's predictive performance.  Given a fitted model, we obtain predictions on the calibration data. Let \(\hat{y}_j\) be the predicted value for observation \(j\) in the calibration set. The residuals, defined as \(r_j = |y_j - \hat{y}_j|\), reflect the differences between the observed and predicted values. These residuals help assess model prediction error, which indirectly reflects uncertainty~\cite{da2024tutorial}.

To establish a confidence interval for a new observation \(\boldsymbol{x}_*\), we first calculate the quantiles of the residuals. Specifically, we compute the \((1-\alpha)\) quantile of the residuals from the calibration set, denoted as \(q_{\alpha}\), where \(\alpha\) represents the significance level. This can be mathematically represented as:
\[
q_{\alpha} = \text{Quantile}_{1-\alpha}(r_1, r_2, \ldots, r_{n_{\text{cal}}}),
\]
where \(r_j\) are the computed residuals from the calibration set. Finally, the confidence interval for a new observation \(\boldsymbol{x}_*\) can be expressed as:
\[
\hat{y}_* \pm q_{\alpha},
\]
where \(\hat{y}_*\) is the prediction made by the model for \(\boldsymbol{x}_*\). This confidence interval guarantees valid coverage probability under the assumption that the calibration set is representative of the underlying distribution of the data. 

Theoretical guarantees for conformal prediction stem from the notion of exchangeability, which ensures that the predictions made on unseen data are valid, meaning they will cover the true values with the desired probability \(1 - \alpha\) \cite{shafer2008tutorial}.  This confidence interval guarantees valid coverage probability under the assumption that the calibration set is representative of the underlying distribution of the data. In the following, we will consider $\alpha = 10\%$, 80\% of the data to train the model and the rest 20\% for model calibration.

SMT does not come with tools to analyze and explain the dependency between variables and outputs, only the previously mentioned basic uncertainty quantification tools, therefore motivating for dedicated tools.

\section{Explainability module for SMT-EX}
SMT-EX incorporates three methods: SHAP, PDP, and ICE. These methods were deliberately chosen due to their widespread use in various machine-learning applications. They are well-suited to our objective of providing insight and knowledge to SMT users who build surrogate models for solving diverse engineering problems. The SMT-EX itself can be used with any type of surrogate model provided by SMT since the only required information is the prediction information. However, for demonstration purposes, we focus on the GPR from SMT since it can also handle mixed-categorical modelling. In addition, SMT-EX also implements Sobol' indices for problems with all continuous variables. 

SHAP assigns each feature an importance value for a particular prediction, offering a comprehensive view of feature impact by considering all possible feature combinations. PDP illustrates the relationship between a feature and the predicted outcome by showing the average effect of a feature while averaging out the influence of other features. Finally, ICE plots provide a detailed view by showing how the prediction changes for individual instances as a particular feature varies, highlighting interactions and heterogeneities in feature effects. This section briefly explains these three methods and Sobol' indices.

In the following explanations, we will use the following notations. First, let $[1:m] :=\{1,\ldots,m\}$ and $U \subseteq [1:m] $ and $C := [1:m] \setminus U$ as the complement of $U$. Further, let $\boldsymbol{x}_U := \{ x_j : j \in U \}$ denote the vector of the input of interest and $\boldsymbol{x}_{C} := \{ x_j : j \in C \}$ be the set of the other output. Let $|.|$ be the cardinality of a set, we are primarily interested in cases with $|U|=1$ or $|U|=2$ when performing visualization or GSA.
\subsection{Sobol' indices}
Sobol' decomposition works according to the following decomposition
\begin{equation}
    f(\boldsymbol{x}) = f_{\emptyset} + \sum_{i=1}^m f_i(x_i) + \sum_{1 \le i < j \le m} f_{i,j}(x_i, x_j) + \cdots + f_{1,2,\ldots,m}(x_1, x_2, \ldots, x_m)
\end{equation}
where $f_{\emptyset}$ is defined such that
\begin{equation}
    f_{\emptyset} = \mathbb{E}[f(\boldsymbol{x})],
\end{equation}
where $\mathbb{E}[.]$ is the expectation operator. In this regard, $f_i$ is the main effect of variable $x_{i}$ and higher cardinality indicates interaction, \textit{e.g.}, $f_{i,j}$ denotes the interaction effect between $x_i$ and $x_j$. Each $f_{U}(\boldsymbol{x}_{U})$ term but $f_{\emptyset}$ satisfies the following property:
\begin{equation}
    \mathbb{E}[f_U(\boldsymbol{x}_U) | x_k] = 0 \quad \text{for any } k \in S.
\end{equation}

The Sobol' indices quantify the contribution of each subset $U$ to the total variance of $f(\boldsymbol{x})$. The total variance is defined as
\begin{equation}
 \mathrm{Var}[f(\boldsymbol{x})] = \sum_{\emptyset \neq U \subseteq [1:m]} \mathrm{Var}[f_U(\boldsymbol{x}_U)]
\end{equation}
where $\mathrm{Var}[f_U(\boldsymbol{x}_U)]$ denotes the partial variance correspond to the subset $U$. We then have the following definitions of Sobol' indices:
\begin{equation}
    S_U = \frac{\mathrm{Var}[f_U(\boldsymbol{x}_U)]}{\mathrm{Var}[f(\boldsymbol{x})]}.
\end{equation}
For example, $S_1$ denotes the main effect of variable $x_{1}$, $S_{1,2}$ denotes the interaction effect due to $x_{1}$ and $x_{2}$.

A related metric is the total Sobol' index which represents the proportion of the output variance that can be attributed to the input variable, including all of its interactions with other variables. The total Sobol' index for the $i$-th variable is defined as 
\begin{equation}
    S_{T_i} = \sum_{\substack{U \subseteq [1:m] \\ i \in U}} S_U.
\end{equation}
For example, if $m=3$ then $S_{T_{1}}$ is defined as $S_{T_{1}}= S_{1} + S_{1,2} + S_{1,3} + S_{1,2,3}$~\cite{sobol2001global}.

A Monte Carlo simulation can be used to estimate Sobol' indices for a given $f(\boldsymbol{x})$ and the range of $\boldsymbol{x}$~\cite{gamboa2016statistical}. It is worth noting that Sobol' indices are only valid in their general formulation for continuous (or at least ordered) variables because they rely on smoothly decomposing output variance across an infinite range of input values~\cite{chan1997sensitivity}. Furthermore, Sobol' indices provide only the importance of input variables and do not reveal the specific ways an input variable influences the output. This limitation necessitates the use of explainability methods like PDP, ICE, and SHAP, which become the core of the current SMT-EX version.

\subsection{Partial Dependence Plots and Individual Conditional Expectations}
The partial dependence function~\cite{friedman2001greedy} can be expressed as
\begin{equation}
    \label{eq:PDP}
    f_{pd}(x_{U}) = \mathbb{E}_{\boldsymbol{x}_{C}}\big[\hat{f}(\boldsymbol{x}_{U},\boldsymbol{x}_{C}) \big] = \int f (\boldsymbol{x}_{U},\boldsymbol{x}_{C}) d\mathrm{P}(\boldsymbol{x}_{C}) \approx \frac{1}{n_{r}} \sum_{i=1}^{n_{r}}f(\boldsymbol{x}_{U},\boldsymbol{x}_{C}^{(i)}),
\end{equation}
where $n_{r}$ is the size of points for the estimation of PDP. Creating a plot of the partial dependence function illustrates how the subset of input variables (\textit{i.e.}, $\boldsymbol{x}_{U}$) affects the model's predictions while accounting for the influence of other features (\textit{i.e.}, $\boldsymbol{x}_{C}$) through marginalization or averaging. Observing the average predictions at varying values of the chosen feature allows us to detect patterns and trends, making the model easier to interpret and aiding informed decision-making. To put it simply, PDP shows the overall trend observed when a specific input changes, helping to identify trends like nonlinearity (in an average sense). However, since PDP reflects an average impact, some details might be obscured due to this aggregation. For example, PDP does not visualize possible interactions due to the interaction between $\boldsymbol{x}_{S}$ and the other variable. 

ICE plots extend PDPs by showing the effect of an input variable on predictions at the individual observation level~\cite{goldstein2015peeking}. Instead of averaging over all data points, an ICE plot provides a line for each instance, revealing how changing an input value affects that specific observation. In consequence, an ICE curve is essentially $\hat{f}(x_{U},\boldsymbol{x}_{C}^{(i)})$ as in Eq.~(\ref{eq:PDP}) for a sample $\boldsymbol{x}^{(i)}$. It is worth noting that PDP itself is essentially an aggregation of ICE curves. By averaging the individual effects, PDP provides a more general view of how an input variable impacts prediction. ICE plots, on the other hand, help identify interactions or non-linear relationships by showing the variation in how different observations respond to changes in a particular input. In practice, both PDP and multiple ICE curves are often depicted together, so that the trend can be seen from both global and more granular viewpoints.

One useful feature of PDP is the feature importance derived from the Partial Dependence (PD) function~\cite{greenwell2018simple}. In this regard, feature importance from PDP calculates the standard deviation of the partial dependence function, reflecting the variability and influence of a specific feature on the model's predictions. Let us denote the corresponding feature importance for the $j$-the variable as $I_{PD}(x_{j})$ (\textit{i.e.}, $|U|=1$):
\begin{equation}
    I_{PD}(x_{j}) = \sqrt{\frac{1}{n_{PD}-1} \sum_{i=1}^{n_{PD}}\big(f_{pd}(x_{j}^{(i)})-\frac{1}{n_{PD}} \sum_{i=1}^{n_{PD}}f_{pd}(x_{j}^{(i)})\big)^{2}},
\end{equation}
where $n_{PD}$ is the number of samples tried for calculating $I_{PD}$. Unlike Sobol' indices, the PD feature importance also applies for categorical variables with some modifications. For categorical inputs, the feature importance is calculated as follows:
\begin{equation}
    I_{PD}(x_{j}) =\left(\text{max} _k\left(f_{pd}\left(x_{s}^{(k)}\right)\right)-\text{min} _k\left(\hat{f}_S\left(x_{s}^{(k)}\right)\right)\right) / 4,
\end{equation}
where $k$ is the number of unique values for the categorical feature of interest.

\subsection{Shapley Additive Explanations and Individual Conditional Expectations}


For easier notations in the following explanations, let us denote $-U = [1:m]\setminus U$. A Shapley value can be computed for each data point, indicating the importance of the associated feature in classifying that specific data point $\boldsymbol{x}^{(i)}$, which reads as follows:
\begin{equation}
\label{eq:Shapley_value}
\phi_j (\boldsymbol{x}^{(i)}) = \frac{1}{m} \sum_{U \subseteq {-j}} \binom{m-1}{|U|}^{-1} (\operatorname{val}(U \cup {j}) - \operatorname{val}(U)),
\end{equation}
where $\text{val}(.)$ denotes the value function, which is based on the model's prediction and measures the benefit gained from individual or multiple variables working together~\cite{shapley1953value}. For example, $\operatorname{val}(U) = \hat{f}(\boldsymbol{x}_{U}^{(i)})$, where $\boldsymbol{x}_{U}^{(i)}$ is the subset of $\boldsymbol{x}^{(i)}$ that takes only the variables involved in $U$. In practice, SHAP is computed using approximation algorithms such as KernelSHAP to accelerate the computation time~\cite{covert2021improving} for high-dimensional input. In SMT-EX, both exact and KernelSHAP are implemented. To calculate SHAP values, a reference value must be defined. For continuous variables with a known domain, it is advisable to set the reference value at the center of the input domain. When categorical variables are involved, a single unique value from the categorical feature should be chosen as the reference.

SHAP values are inherently localized, applying to individual samples within the input space. However, engineers often seek to understand the global impact of a specific input variable on the QoI. By plotting SHAP values across multiple samples in the input space, engineers can gain insights into how a particular input variable influences the QoI in a broader sense. This SHAP dependence plot allows for the observation of interactions and nonlinearities, providing a clearer overall picture. One useful feature of SHAP is that it provides the means for global sensitivity analysis through the averaged SHAP values. The averaged 
 SHAP value for the $j$-th variable ($|\phi_{j}|$), can be computed as follows:
\begin{equation}
    |\phi_{j}| = \frac{1}{n_{sh}}\sum_{i=1}^{n_{sh}}|\phi_{j}(\boldsymbol{x})^{(i)}|
\end{equation}
where $n_{sh}$ is the number of samples used for the calculation of averaged SHAP values.

Both PDP/ICE and SHAP can be used for mixed-categorical problems, which makes them useful for general applications in which non-continuous variables are also present.

\section{Application to engineering problems}

\subsection{A continuous 10-variable problem: Wing weight function}
The first engineering application studied here is the wing weight problem that involves an analytical function that represents a light aircraft wing~\cite{forrester2008engineering}. The target output is the wing's weight. The formula for calculating the wing weight is as follows:
\begin{equation}
		\label{eq:ww}
		f(\boldsymbol{x}) = 0.036S_{w}^{0.758} W_{fw}^{0.0035} \bigg(\frac{A}{cos^{2}(\Delta)}\bigg)^{0.6} q^{0.006} \lambda^{0.04} \bigg(\frac{100t_{c}}{cos(\Delta)}\bigg)^{-0.3} (N_{z} W_{dg})^{0.49} + S_{w} W_{p}.
	\end{equation}
Table~\ref{tab:wwdv} shows a nomenclature of the symbols used in Eq.~(\ref{eq:ww}), as well as a baseline set of values, roughly representative of a Cessna C172 Skyhawk aircraft and a somewhat arbitrarily chosen range for each variable~\cite{forrester2008engineering}. The expression shown in Eq.~(\ref{eq:ww}) is quite convoluted and involves trigonometric, power terms, and multiplication of multiple variables, which gives an impression of nonlinearity. Thus, although the model itself is transparent, it is not a useful interpretable expression. We then applied SHAP and PDP/ICE from the SMT-EX module to visualize the behaviour of the 10-variable wing weight problem.

\begin{table}[H]
		\caption{Variables used in the wing weight function.} \label{tab:wwdv}
		\begin{center}
			\begin{tabular}{ | c | c | c |}
				\hline
				\textbf{Variables} & \textbf{Variable Name (unit)} & \textbf{[lower, upper bound]} \\
				\hline
				\(S_{w}\) & Wing area (ft\textsuperscript{2}) & [150, 200] \\ 
				\hline
				\(W_{fw}\) & Weight of fuel in the wing (lb) & [220, 300] \\ 
				\hline
				\(A\) & Aspect ratio & [6, 10] \\ 
				\hline
				\(\Delta\) & Quarter-chord sweep (degrees) & [-10, 10] \\
				\hline
				\(q\) & Dynamic pressure at cruise (lb/ft\textsuperscript{2}) & [16, 45] \\
				\hline
				\(\lambda\) & Taper ratio & [0.5, 1] \\
				\hline
				\(t_{c}\) & Airfoil thickness to chord ratio & [0.08, 0.18] \\
				\hline 
				\(N_{z}\) & Ultimate load factor & [2.5, 6] \\
				\hline 
				\(W_{dg}\) & Flight design gross weight (lb) & [1700, 2500] \\
				\hline
				\(W_{p}\) & Paint weight (lb/ft\textsuperscript{2}) & [0.025, 0.08] \\
				\hline
			\end{tabular}
		\end{center}
	\end{table}

\begin{figure}[H]
    \centering
    \includegraphics[width=0.9\linewidth]{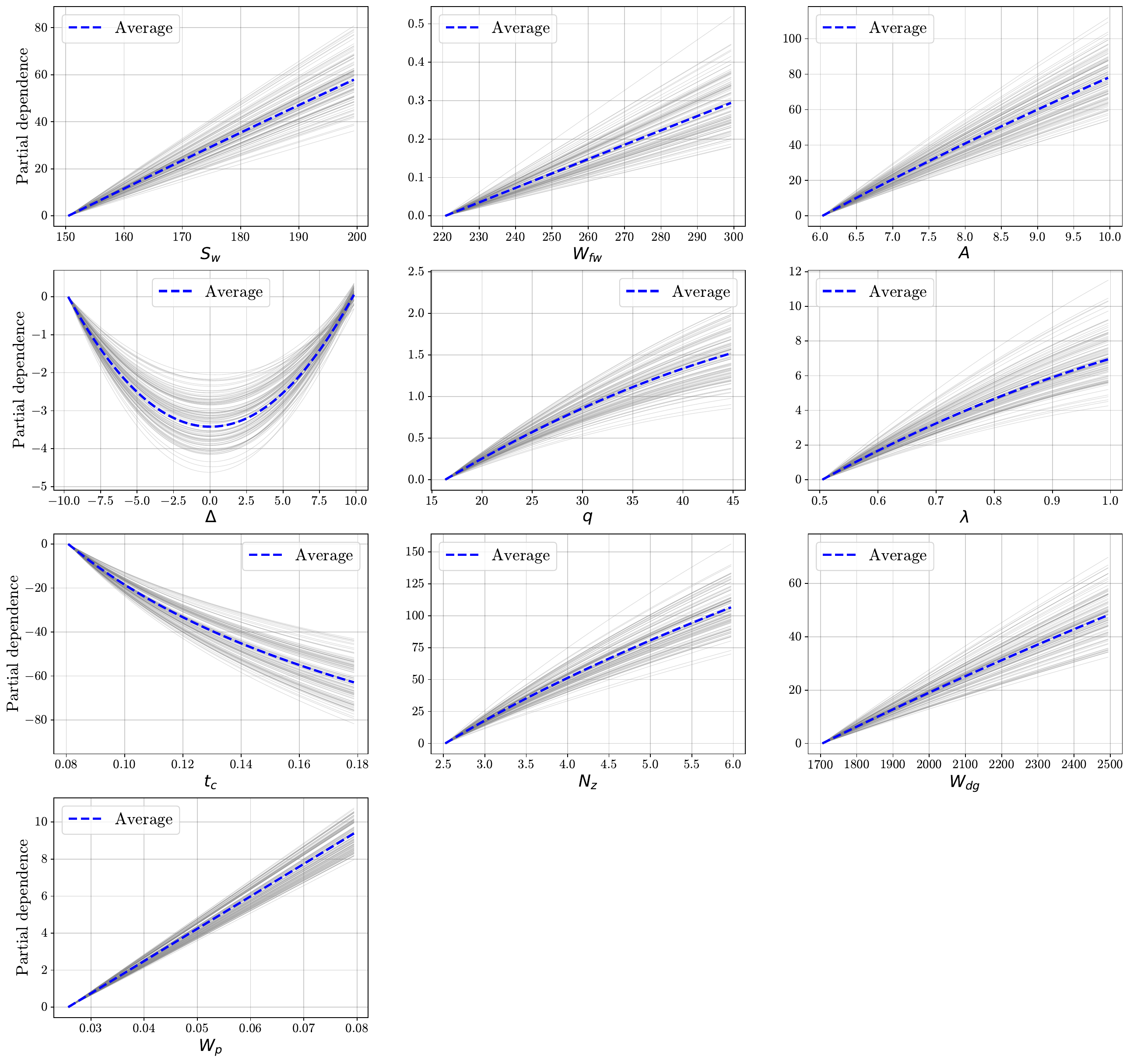}
    \caption{Partial dependence plots and individual conditional expectations for the wing weight problem.}
    \label{fig:wing_weight_pdp}
\end{figure}

A GPR with a squared exponential covariance function was used as a surrogate model for this problem. For this problem, 300 data points were initially generated from the true function, Eq.(\ref{eq:ww}), where 80\% of the data is used to train the model and the rest 20\% for model evaluation. After confirming its decent predictive performance that yields RMSE=0.144 in the test data, we employ SMT-EX to gain an explanation of the model.   The SHAP and PDP/ICE plots generated by SMT-EX are shown in Fig.~\ref{fig:wing_weight_pdp}. Essentially, PDPs show that the trends are monotonous to all variables but the quarter chord sweep of the wing that alters the weight in a quadratic fashion. Although this makes sense from an engineering viewpoint, it is not easy to quickly grasp this from looking at Eq.~(\ref{eq:ww}). In this regard, explainability provides a valuable tool for examining the impact of each input variable on the output. Another information that can be deduced from the PDPs is how each input variable associates with the output. In addition to the quadratic trend resulting from the quarter chord sweep, it is evident that increasing all variables except  $t_{c}$ generally leads to an increase in wing weight. This insight is particularly valuable in applications like optimization, where understanding how to adjust input variables to optimize the objective function is essential.

The deviation of ICE curves from the main PDP plot signifies interactions between the variable of interest and other variables. This means that, depending on the values of other variables, the trend shown by the ICE curve might change. Essentially, ICE curves represent one-dimensional slices of the predictive model. Take the effect of $N_{z}$ as an example. It can be observed that the gradient of each ICE curve differs, indicating that the impact of the slope on $N_{z}$ varies based on the values of other input variables. Next, let us now examine the quarter-chord sweep ($\Delta$), which consistently exhibits a quadratic trend regardless of the values of the other inputs, although its relative impact varies. ICE curves are also valuable for capturing changes in the association between an input variable and the output (\textit{e.g.}, a shift from positive to negative correlation), though this characteristic is not observed in the current problem.

\begin{figure}[H]
    \centering
    \includegraphics[width=0.9\linewidth]{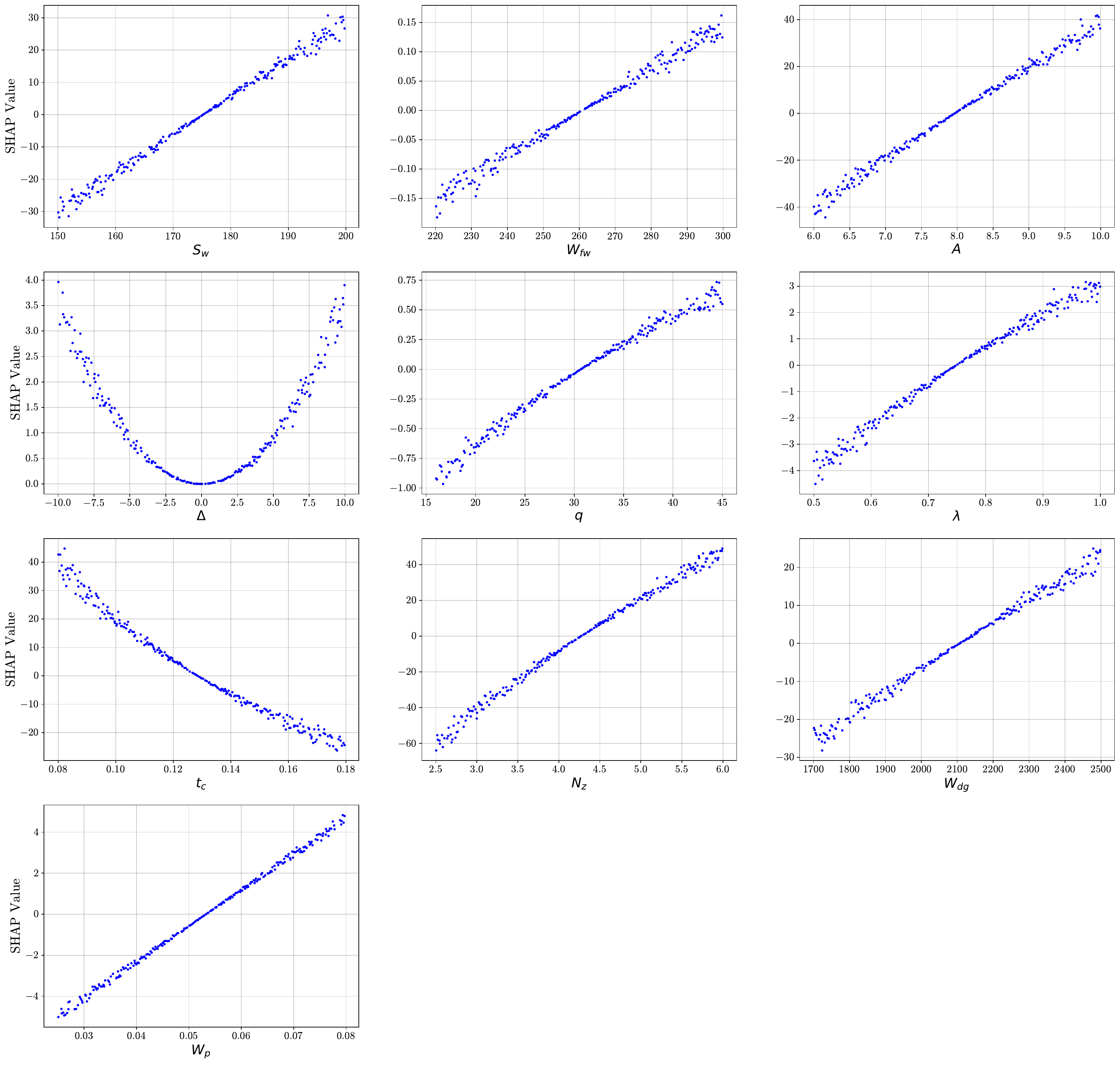}
    \caption{SHAP dependence plots for the wing weight problem.}
    \label{fig:wing_weight_shap}
\end{figure}

The SHAP dependence plots for the wing weight problem are presented in Fig.~\ref{fig:wing_weight_shap}. Both plots (\textit{i.e.}, SHAP and PDP/ICE) essentially convey similar information regarding the trend and magnitude of the impact. However, SHAP offers a different perspective on interactions compared to PDP and ICE. While PDP and ICE plots make it challenging to assess the strength of interactions, they provide a clearer view of the interaction dynamics. On the other hand, SHAP clearly demonstrates the presence of interactions, as indicated by the dispersion of SHAP values (if no interaction existed, there would be no dispersion). The interaction level is shown to be relatively weak here, as the dispersions are small, aligning with the expected intuition for the wing weight problem. Despite that, it is not easy to comprehend the impact of interactions on significantly less influential variables. For example, it is not easy to understand the interaction dynamics for $q$ and $W_{fw}$. It is worth noting that visualization can also be shown by colouring the dots according to the magnitude of the other input variable.

\begin{figure}[H]
    \centering
    \subfigure[Based on PDP]{
    \includegraphics[width=0.45\linewidth]{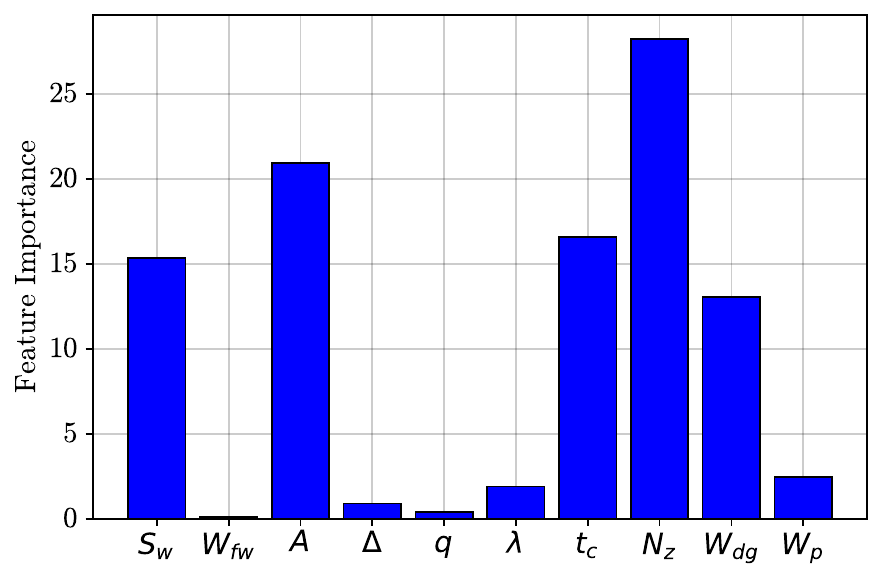}
    }
    \subfigure[Based on SHAP Value]{
    \includegraphics[width=0.45\linewidth]{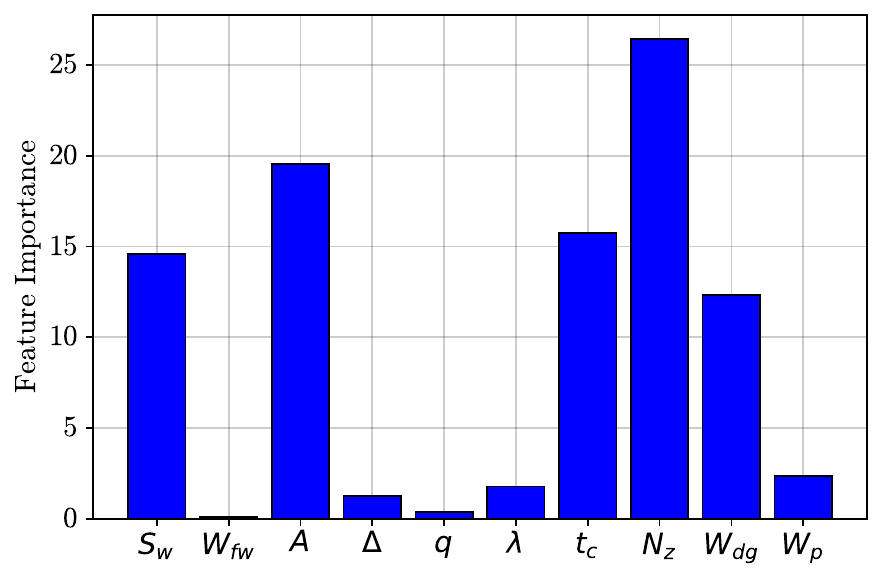}
    }
    \caption{Feature importance of the wing weight problem based on PDP and SHAP.}
    \label{fig:wing_weight_feature_importance}
\end{figure}

\begin{figure}[H]
    \centering
    \subfigure[First order Sobol' indices]{
    \includegraphics[width=0.45\linewidth]{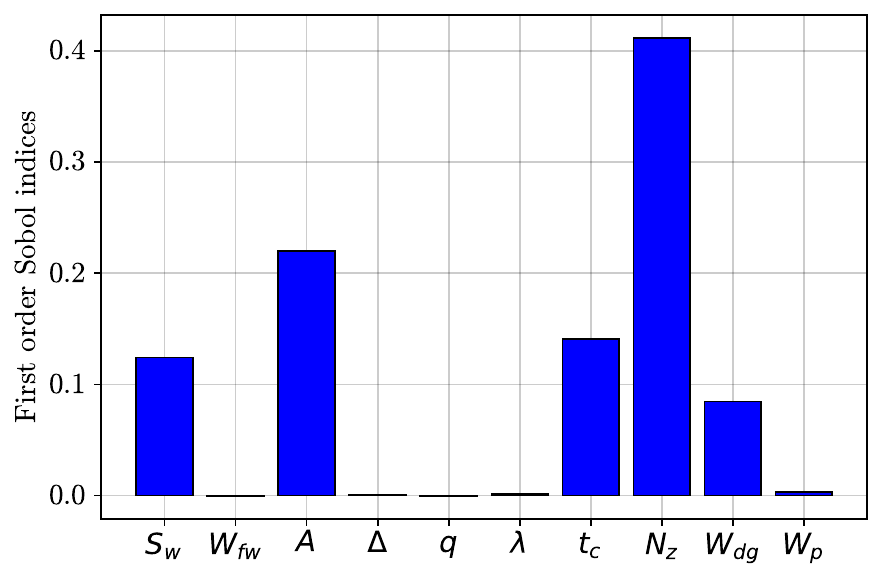}
    }
    \subfigure[Total order Sobol' indices]{
    \includegraphics[width=0.45\linewidth]{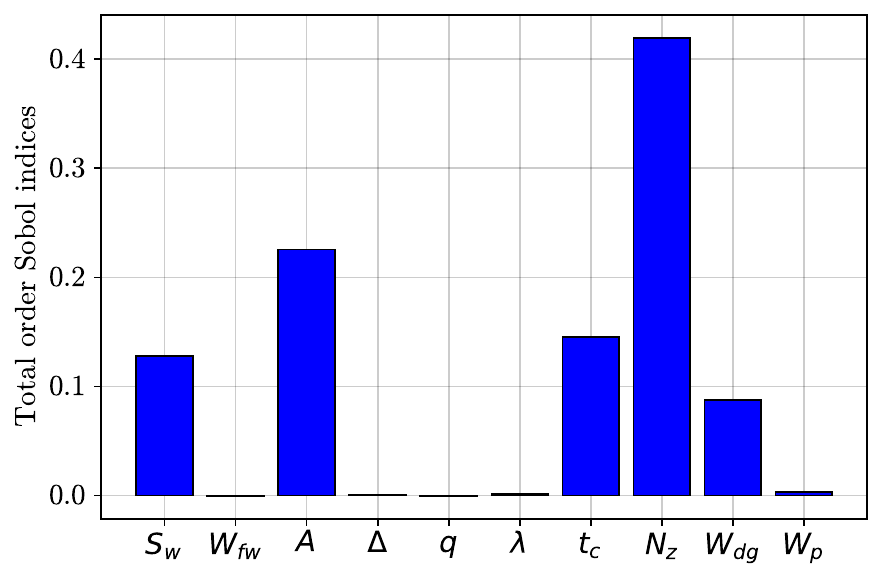}
    }
    \caption{First and total order Sobol' indices for the wing weight problem.}
    \label{fig:wing_weight_sobol_indices}
\end{figure}

Finally, the input importance plots (\textit{i.e.}, GSA) for the wing weight problems according to PDP and SHAP generated by SMT-EX are shown in Fig.~\ref{fig:wing_weight_feature_importance}. Besides, we also show the first and total order Sobol' indices in Fig.~\ref{fig:wing_weight_sobol_indices}, a feature that is also implemented in SMT-EX. All feature importance metrics convey similar information, in which $N_{z}$ is the most influential variable on the wing weight, followed by $A$ and $t_{c}$. Interestingly, it can be seen that the magnitudes of the feature importance from PDP and SHAP are similar, despite the different principles used to measure the feature importance. The two methods also successfully detect that some variables are non-influential, \textit{e.g.}, $\lambda$, and $W_{fw}$. The observation aligns with the results of Morris's method presented in our previous paper~\cite{palar2023shapley} and Sobol' indices. It is worth noting that the feature importance from PDP measures the standard deviation of the partial dependence function, indicating the variability and influence of specific input variables on the model's predictions, on average. On the other hand, the averaged SHAP quantifies the contribution of each feature to the model's predictions by averaging the absolute Shapley values of that feature across all samples. It is also important to note that the relative differences between the Sobol' indices of all features are noticeably larger compared to the feature importance values from PDP and SHAP. This is because Sobol' indices are derived from variance and are squared quantities, which amplifies the differences.

This demonstration highlights the capabilities of SMT-EX in explaining the complexity of problems involving continuous variables. It is important to emphasize that while Sobol' indices quantify feature importance, they do not reveal the internal workings of the black-box function. In contrast, SHAP and PDP/ICE provide insights into these dynamics. As a surrogate modeling tool, SMT stands out by offering not only feature importance and GSA but also a deeper understanding of how input variables influence the predictions- which is enabled thanks to the SMT-EX.

\begin{figure}[H]
    \centering
    \hspace{-1cm}
\includegraphics[height=14cm,width=1.05\linewidth]{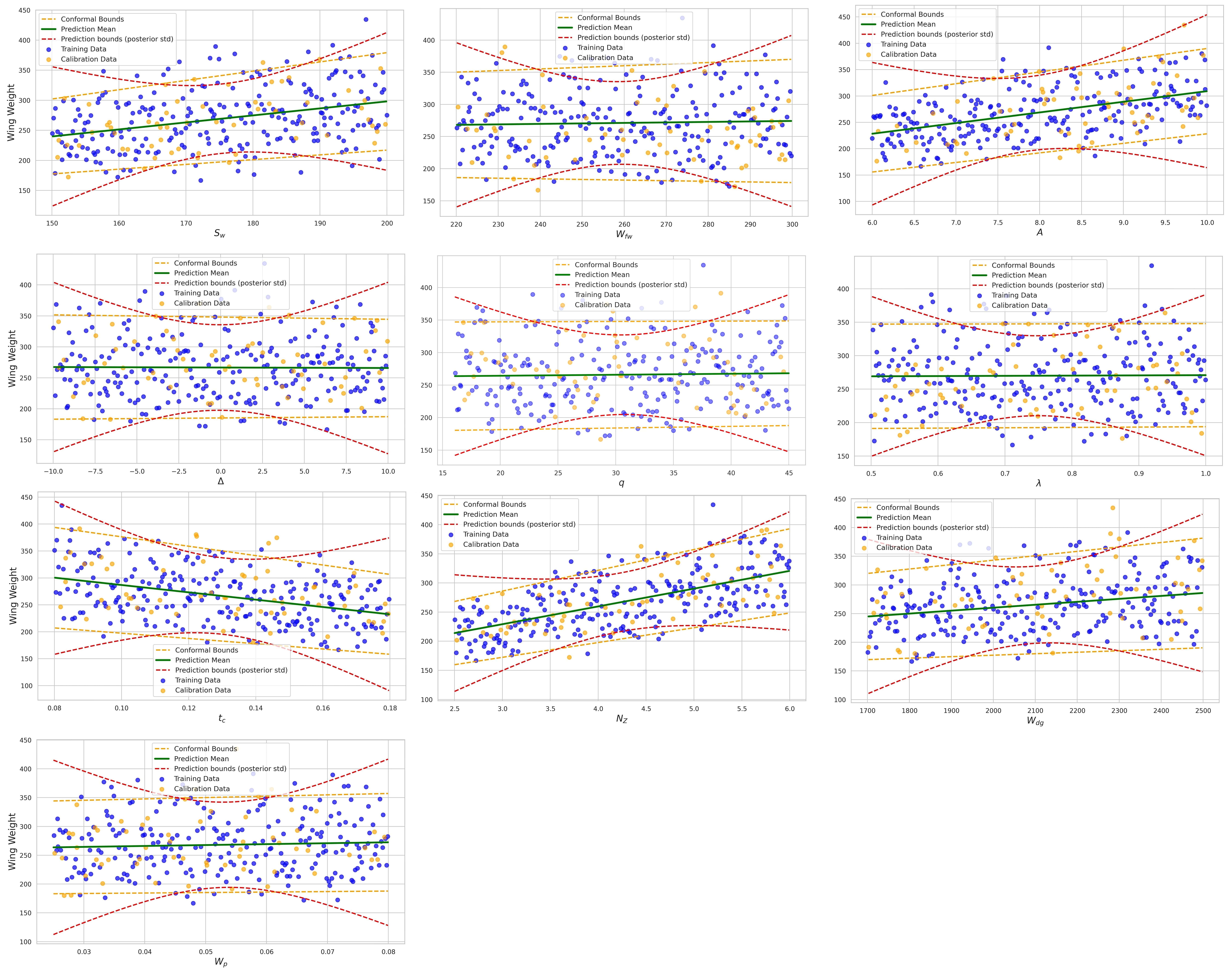}
    \caption{Prediction intervals for the wing weight problem.}
    \label{fig:wing_conf}
\end{figure}

The confidence intervals are displayed in Fig.~\ref{fig:wing_conf}. Notice that a single GP was built with a single variable one at a time to provide insight into how conformal prediction performs compared to uncertainties from GP.  To compute them, we use the same training and validation sets generated by LHS to cover the whole space. Using a surrogate model, such as a GP, we can see how moving along one variable affect the data dispersion on the remaining variables and quantify the dispersion. We can already make some comments on the results. Indeed, the conformal predictions are better than the confidence intervals as they are more adaptive to the position of a given point but here the uncertainty is not highly informative, the models are linear and the conformal intervals are similarly linear. The predicted uncertainty is also higher on the boundary of the domain which is a known effect for when the boundary lacks points.  More importantly, the variables $W_{fw}$, $\Delta$, q, $\lambda$ and $W_p$ seem to have almost no effect on the wing weight output, confirming the feature importance analysis and going in the same sense as the Sobol' indices interpretation. Obviously, the bigger the aircraft, the more its weight and therefore increasing wing area $S_W$ or wing aspect ratio $A$ increases the output. Similarly, the wing should be more massive when the load factor $N_Z$ and the design weight $W_{dg}$ increase. To finish with, the more the thickness to chord ratio $t_c$, the less the wing weight, probably because high values correspond to low aspect ratio. Compared to the previous one, this analysis can quantify the effect of a variable on the output along the range (positive or negative effect/linearity,...). Moreover, it is consistent with the feature importance as the model for $N_Z$ explains more the variance and, as such,  yields to the tighter confidence intervals.

\subsection{ A mixed-categorical problem: Cantilever beam bending}
\begin{figure}[H]
\centering
\captionsetup{justification=raggedright,singlelinecheck=false}

\begin{minipage}[t]{0.8\textwidth}
\hspace{-5.25cm}

\begin{tikzpicture}
    \point{origin}{-0.75}{-0.25};
    \point{begin}{0}{0};
    \point{end}{5}{0};
    \point{end_bot}{4.99}{-0.9};
    \point{end_up}{5}{0.5};
    \beam{2}{begin}{end};
    \support{3}{begin}[-90];
    \load{1}{end}[90]   ;
    \notation{1}{end_up}{$F=50kN$};

     \draw[<->] (end) -- (end_bot) node[midway, right] {$\delta$} ;
     \draw[<->] (0,0.5) -- (5,0.5) node[midway, above] {L};
     
    \draw
      [-, ultra thick] (begin) .. controls (1.5, +.01) and (2.5, -.15) .. (4.93, -0.9)
      [-, ultra thick] (begin) .. controls (1.5, +.01) and (2.5, -.2) .. (4.85, -1.5)
      [-, ultra thick] (begin) .. controls (1.5, +.01) and (2.5, -.4)   .. (4.78, -1.9);
  \end{tikzpicture}
    \end{minipage}%

\vspace{-3cm}
\hfill
\hspace{-5.25cm}
\begin{minipage}[t]{0.45\textwidth}
\centering
\begin{tikzpicture}


\tstar{0.25}{0.5}{6}{0}{thick,fill=yellow,xshift=+3.6cm,yshift= -1.2cm}
\tstar{0.14}{0.28}{6}{0}{thick,fill=white,xshift=+3.6cm,yshift= -1.2cm}

\tstar{0.25}{0.5}{6}{0}{thick,fill=yellow,xshift=+2.4cm,yshift= -1.2cm}
\tstar{0.08}{0.16}{6}{0}{thick,fill=white,xshift=+2.4cm,yshift= -1.2cm}

\tstar{0.25}{0.5}{6}{0}{thick,fill=yellow,xshift=+1.2cm,yshift= -1.2cm}

\fill[green,even odd rule] (3.6,0) circle (0.5) (3.6,0) circle (0.33);
\draw (3.6,0) circle (0.5) ;
\draw (3.6,0) circle (0.33) 
; 
\fill[green,even odd rule] (2.4,0) circle (0.5)(2.4,0) circle (0.17);
\draw (2.4,0) circle (0.5) ;
\draw (2.4,0) circle (0.17) 
; 
\fill[green,even odd rule] (1.2,0) circle (0.5) ;
\draw (1.2,0) circle (0.5) ;

\def\pos{-2.4}
\fill[blue,even odd rule]  (\pos-0.5,-1.7+1.2) -- (\pos-0.5,-0.7+1.2) -- (\pos+0.5,-0.7+1.2) -- (\pos+0.5,-1.7+1.2) -- cycle ;

\def\pos{-1.2}
\fill[blue,even odd rule]  (\pos-0.5,-1.7+1.2) -- (\pos-0.5,-0.7+1.2) -- (\pos+0.5,-0.7+1.2) -- (\pos+0.5,-1.7+1.2) -- cycle   (\pos-0.25,-1.45+1.2) -- (\pos-0.25,-0.95+1.2) -- (\pos+0.25,-0.95+1.2) -- (\pos+0.25,-1.45+1.2) -- cycle ;

\def\pos{0}
\fill[blue,even odd rule]  (\pos-0.5,-1.7+1.2) -- (\pos-0.5,-0.7+1.2) -- (\pos+0.5,-0.7+1.2) -- (\pos+0.5,-1.7+1.2) -- cycle   (\pos-0.35,-1.55+1.2) -- (\pos-0.35,-0.85+1.2) -- (\pos+0.35,-0.85+1.2) -- (\pos+0.35,-1.55+1.2) -- cycle ;

\def\pos{-2.4}
\fill[black] (\pos-0.24,-0.5-1.2) -- (\pos-0.24,0.5-1.2) -- (\pos+0.24,0.5-1.2)  -- (\pos+0.24,-0.5-1.2)   -- cycle ;
\fill[black] (\pos-0.5,-0.5-1.2) -- (\pos-0.5,-0.18-1.2) -- (\pos+0.5,-0.18-1.2)  -- (\pos+0.5,-0.5-1.2)   -- cycle ; 
\fill[black] (\pos-0.5,0.18-1.2) -- (\pos-0.5,0.5-1.2) -- (\pos+0.5,0.5-1.2)  -- (\pos+0.5,0.18-1.2)   -- cycle ;  

\def\pos{-1.2}
\fill[black] (\pos-0.19,-0.5-1.2) -- (\pos-0.19,0.5-1.2) -- (\pos+0.19,0.5-1.2)  -- (\pos+0.19,-0.5-1.2)   -- cycle ;
\fill[black] (\pos-0.5,-0.5-1.2) -- (\pos-0.5,-0.25-1.2) -- (\pos+0.5,-0.25-1.2)  -- (\pos+0.5,-0.5-1.2)   -- cycle ; 
\fill[black] (\pos-0.5,0.25-1.2) -- (\pos-0.5,0.5-1.2) -- (\pos+0.5,0.5-1.2)  -- (\pos+0.5,0.25-1.2)   -- cycle ;  

\def\pos{0}
\fill[black] (\pos-0.14,-0.5-1.2) -- (\pos-0.14,0.5-1.2) -- (\pos+0.14,0.5-1.2)  -- (\pos+0.14,-0.5-1.2)   -- cycle ;
\fill[black] (\pos-0.5,-0.5-1.2) -- (\pos-0.5,-0.32-1.2) -- (\pos+0.5,-0.32-1.2)  -- (\pos+0.5,-0.5-1.2)   -- cycle ; 
\fill[black] (\pos-0.5,0.32-1.2) -- (\pos-0.5,0.5-1.2) -- (\pos+0.5,0.5-1.2)  -- (\pos+0.5,0.32-1.2)   -- cycle ;  

\point{un}{-2.15}{-0.8};
\notation{1}{un}{\tiny 1};
\point{deux}{-2.15+1.2}{-0.8};
\notation{1}{deux}{\tiny 2};
\point{trois}{-2.15+2.4}{-0.8};
\notation{1}{trois}{\tiny 3};

\point{quatre}{-2.15+3.6}{-0.8};
\notation{1}{quatre}{\tiny 4};
\point{cinq}{-2.15+4.8}{-0.8};
\notation{1}{cinq}{\tiny 5};
\point{six}{-2.15+6}{-0.8};
\notation{1}{six}{\tiny 6};

\point{sept}{-2.15}{-0.8-1.2};
\notation{1}{sept}{\tiny 7};
\point{huit}{-2.15+1.2}{-0.8-1.2};
\notation{1}{huit}{\tiny 8};
\point{neuf}{-2.15+2.4}{-0.8-1.2};
\notation{1}{neuf}{\tiny 9};

\point{dix}{-2.15+3.6}{-0.8-1.2};
\notation{1}{dix}{\tiny 10};
\point{onze}{-2.15+4.8}{-0.8-1.2};
\notation{1}{onze}{\tiny 11};
\point{douze}{-2.15+6}{-0.8-1.2};
\notation{1}{douze}{\tiny 12};

  \end{tikzpicture}

\end{minipage}
\captionsetup{justification=centering,singlelinecheck=false}
\caption{Cantilever beam problem~\cite[Figure 6]{saves_mixed}.}
\label{fig:beam}
\end{figure}
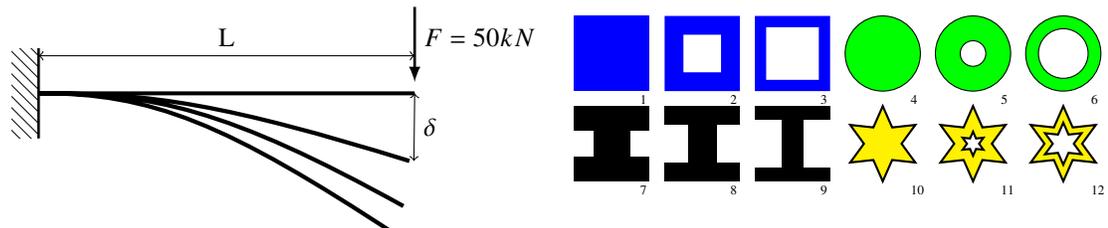

The second engineering application is a beam bending problem in the linear elasticity range that features continuous and categorical variables, as shown in Fig.~\ref{fig:beam}.

The beam is subjected to a force $F$ applied to the free end. The Young modulus and the load were set to $E=200$ GPa and $F=50$  kN, respectively. The problem features two continuous variables, namely, the length $L \in [10,20]$ (in $m$) and the surface $S \in[1,2]$ (in $m^{2}$). Meanwhile, the categorical variable is the type of cross-section with 12 levels~\cite{roustant2020group}, each associated with the normalized moment of inertia $\tilde{I}$ about the neutral axis. The values of the normalized moment of inertia for each cross-section type are described in Table~\ref{tab:categorical_beam}. For the square-shaped cross-section, B is more hollow than A, and C is more hollow than B. This pattern is consistent for the other shapes as well. The output of interest is the tip deflection at the free end, which reads as
\begin{equation}
    \label{eq:cantilever}
    \delta = f(L,S,\tilde{I}) = \frac{F}{3E} \frac{L^{3}}{S^{2}\tilde{I}}.
\end{equation}

\begin{table}[htb]
    \caption{Normalized moment of inertia for each type of cross-section. } \label{tab:categorical_beam}
    \begin{center}
        \begin{tabular}{ | c | c |c|}
            \hline
            \textbf{Cross-section type} & \textbf{Normalized moment of inertia} & \textbf{Shape} \\
            \hline
            A & 0.0833 & Square \\ 
            \hline
            B & 0.139 & Square \\ 
            \hline
            C & 0.380 & Square  \\ 
            \hline
            D & 0.0796& Circle \\ 
            \hline
            E & 0.133 & Circle \\ 
            \hline
            F & 0.363 & Circle \\ 
            \hline
            G & 0.0859 & I-beam \\ 
            \hline
            H & 0.136& I-beam \\ 
            \hline
            I & 0.360  & I-beam \\ 
            \hline
            J & 0.0922 & Star \\ 
            \hline
            K & 0.138  & Star \\ 
            \hline
            L & 0.369 & Star \\ 
            \hline
        \end{tabular}
    \end{center}
\end{table}

Therefore, a mixed-categorical GPR model $\hat{f}(L,S,\tilde{I})$ was built to approximate $f(L,S,\tilde{I})$. To the end, 300 data points are generated from the true function Eq.~(\ref{eq:cantilever}) where we used the same configuration of 80\% samples as the training data and the remaining 20\% for testing. The model obtained an RMSE of $10^{-4}$ in the test data, and we utilized SMT-EX again to extract the behaviour explanations learned by the model.

\textcolor{black}{We first explored the application of conformal prediction for uncertainty quantification in this problem using the mixed-categorical GPR model $\hat{f}(L,S,\tilde{I})$. The boxplots illustrate the true values, the mean predictions from the Gaussian Process (GP), confidence intervals derived from conformal prediction ("conformal lower" and "conformal upper"), and confidence intervals obtained directly from GP ("prediction lower" and "prediction upper") at 300 new validation input points. A key observation here is that the lower bounds from conformal prediction align better with physical expectations compared to those from GP. Specifically, tip deflection should never be negative, as a negative prediction lacks physical validity. In this context, the lower bounds provided by conformal prediction are more reasonable, as they result in significantly fewer instances of negative tip deflection compared to GP.}

\begin{figure}[H]
    \centering
    \hspace{-1cm}
\includegraphics[height=4.5cm,width=1.05\linewidth]{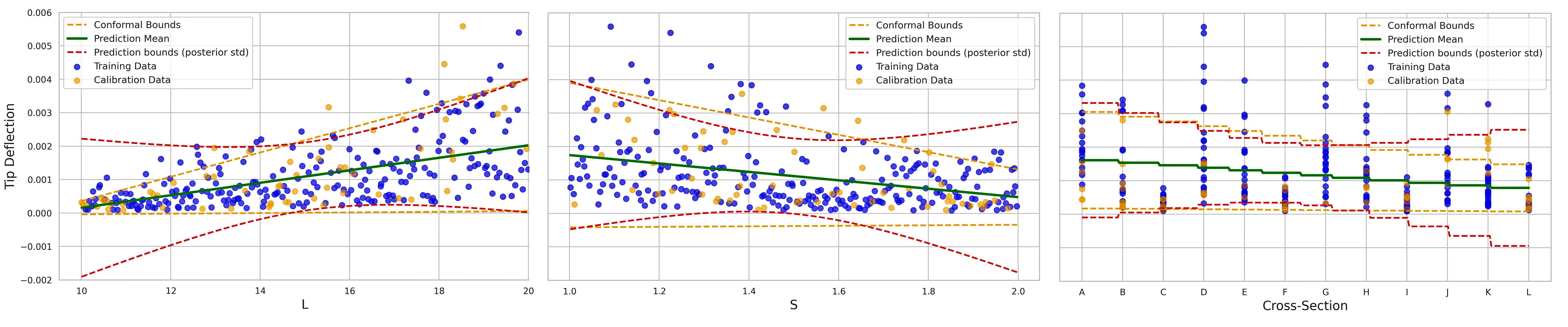}
    \caption{Confidence interval for the cantilever beam problem.}
    \label{fig:cantilever_conf}
\end{figure}

\textcolor{black}{To better understand the confidence intervals provided by conformal prediction, an additional set of GP models was constructed, each considering only one input variable at a time. For instance, a single GP model was developed using only $L$ without considering $S$ and the type of cross-section.} The confidence intervals are displayed in Fig.~\ref{fig:cantilever_conf}. To compute them, we use the same training and validation sets generated by LHS to cover the whole space. Using a surrogate model, such as a GP, we can see how moving along one variable affects the data dispersion on the remaining variables and quantify the dispersion. We can already make some comments on the results. Indeed, the conformal predictions are better than the GP's confidence intervals as they are more adaptive to the position of a given point. More importantly, the lower bound is always between 0 and -0.0004 and always flat. Knowing that the true lower bound is a zero deflection because the response is strictly positive, conformal prediction is better to predict it on all three variables, especially for $L$ where the lower bound is almost perfectly $0$. Moreover, the conformal predictions indicate an increase in uncertainty when $L$ increases and a decrease in uncertainty when $S$ decreases. There is no significant effect of the cross-section shape on the uncertainty. However, more points lead to a decrease in standard deviation which explains why the traditional method underestimates the uncertainty on the cross-section G. Note that most data-centric methods depend highly on the data. For explainability, the initial distribution of points can significantly impact the results. When dealing with synthetic data, choosing the data sample may be a whole field of research to investigate~\cite{bertrand2022variance}. 

\begin{figure}[H]
    \centering
    \includegraphics[width=0.9\linewidth]{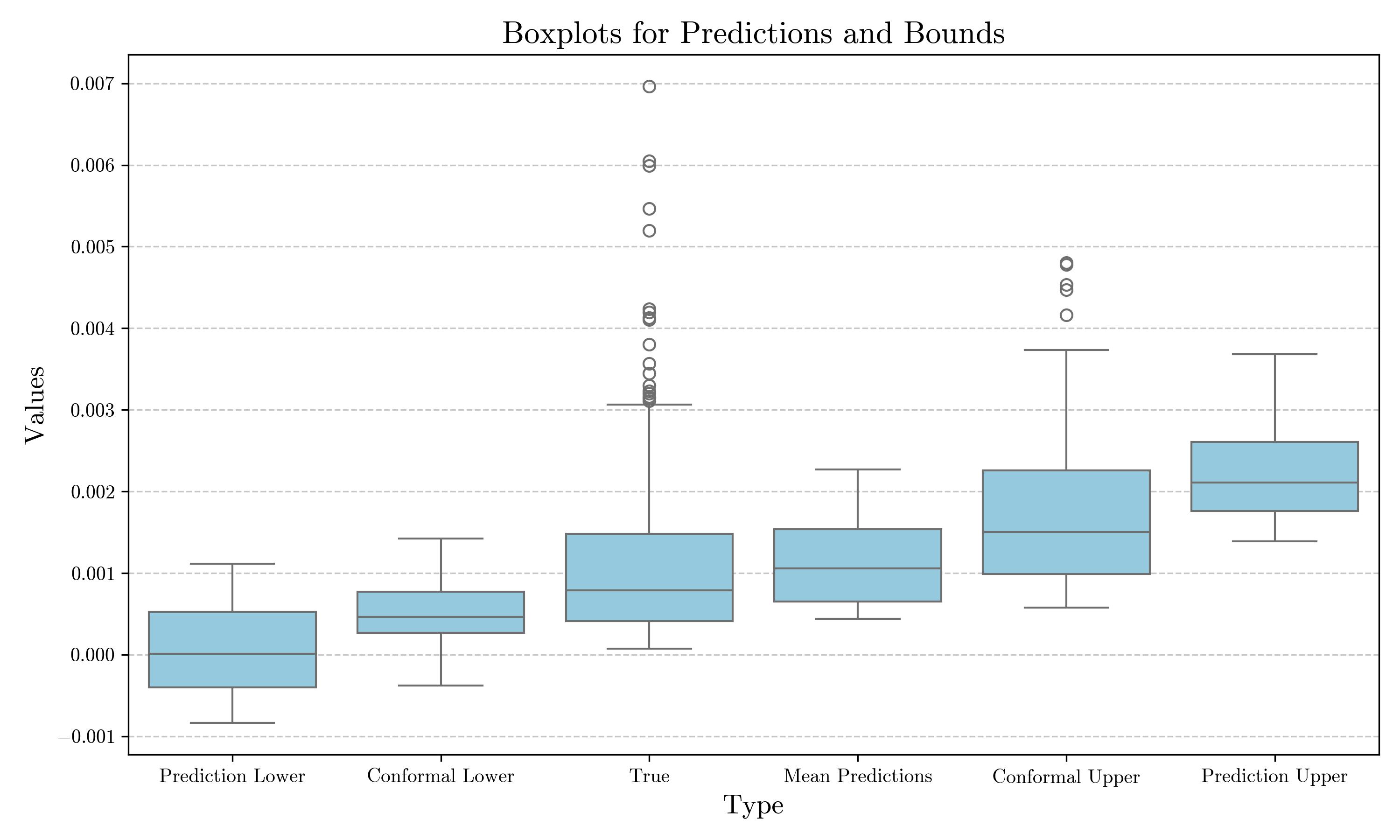}
    \caption{Prediction intervals from the mixed-categorical GPR model $\hat{f}(L,S,\tilde{I})$ for the cantilever beam problem.}
    \label{fig:boxplots2}
\end{figure}

\begin{figure}[H]
\begin{center}
    \begin{minipage}[t]{0.27\textwidth}
        \centering
        \includegraphics[clip=true, height=4.0cm, width=4.2cm]{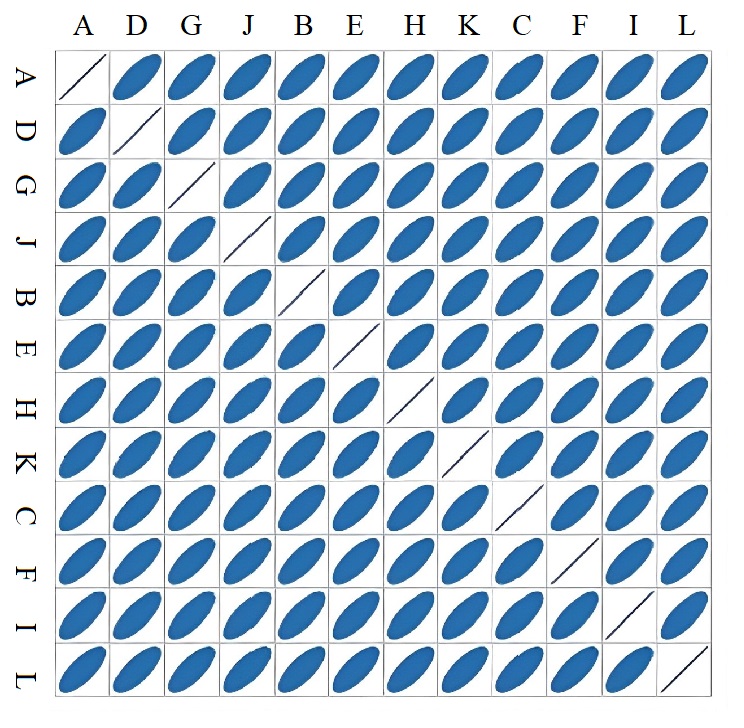}
        \caption*{(a) With GD kernel.}
        \label{corr_canti_gower}
    \end{minipage}%
    \hfill
    \begin{minipage}[t]{0.27\textwidth}
        \centering
        \includegraphics[clip=true, height=4.0cm, width=4.2cm]{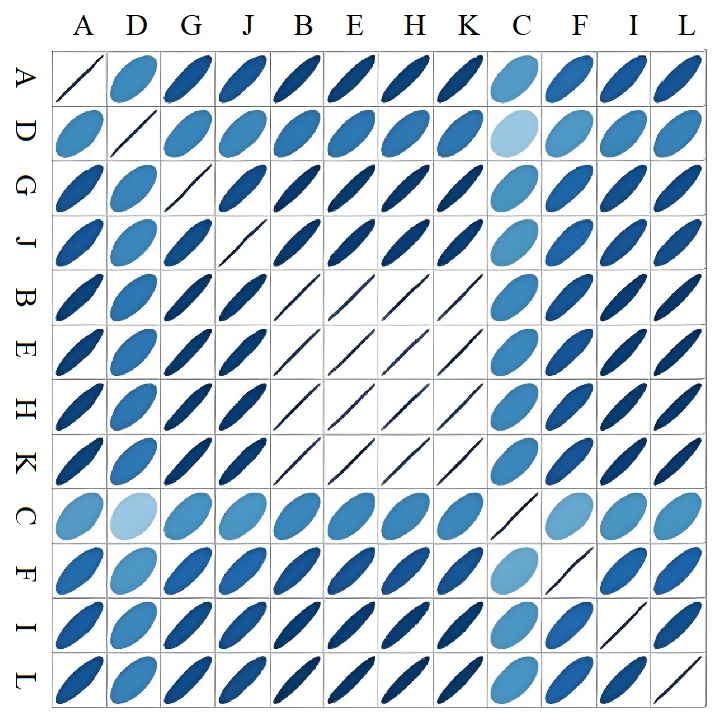}
        \caption*{(b) With CR kernel.}
        \label{corr_canti_cr}
    \end{minipage}%
    \hfill
    \begin{minipage}[t]{0.27\textwidth}
        \centering
        \includegraphics[clip=true, height=4.0cm, width=4.2cm]{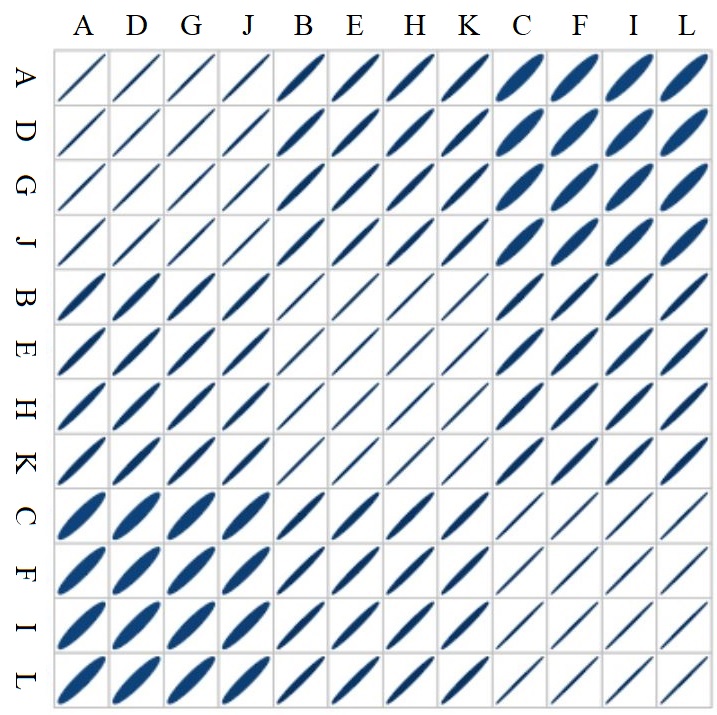}
        \caption*{(c) With EHH kernel.}
        \label{corr_canti_ehh}
    \end{minipage}
        \begin{minipage}[t]{0.15\textwidth}
    \includegraphics[clip=true, height=4.25cm, width=0.5cm]{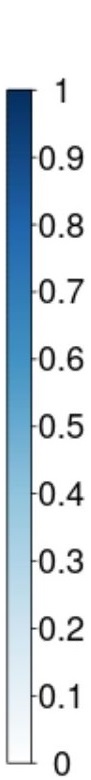}
    \end{minipage}

    \caption{Correlation matrix $R_1^{cat}$ using different choices for $\Theta_1$ for the categorical variable $\tilde{I}$ from the cantilever beam problem.}
    \label{corr_Cantilever}
\end{center}
\end{figure}

Still using the GPR only, in Fig.~\ref{corr_Cantilever}, we have drawn the correlation matrix found between the cross-section shape (the resulting $R_1$ correlation matrix) for  three categorical models: Gower distance, continuous relaxation and homoscedastic hypersphere~\cite{saves_mixed}.  On the figure below, the higher the correlation, the thinner the ellipse.    
As anticipated, the shapes are organized into three distinct groups based on their thickness levels: \{1,4,7,10\}, \{2,5,8,11\}, and \{3,6,9,12\} meaning that the thickness of the cross-section matters more that its shape for predicting the tip deflection. Shapes within the same thickness group exhibit stronger correlations, indicating that thickness has a greater influence on tip deflection than cross-sectional shape. Additionally, in this dataset, any two points with similar values for $L$ and $S$ will yield similar outputs regardless of their cross-sectional design. This consistent influence of cross-section, modelled as $\frac{1}{\tilde{I}}$, results in high correlations when the likelihood is maximized. In Fig.~\ref{corr_Cantilever}c, using the EHH kernel, these three thickness groups can be clearly observed, with correlations nearing 1. Consequently, the homoscedastic hypersphere model~\cite{Pelamatti} would produce a similar correlation matrix. Moreover, with the CR kernel shown in Fig.~\ref{corr_Cantilever}b, the middle-thickness group \{2,5,8,11\} correlates with both the full and hollow groups, resulting in higher correlation values, whereas the correlation hyperparameters associated with the other two groups are smaller. In the GD model (Fig.~\ref{corr_Cantilever}a), only one primary positive correlation value is observed.

\begin{figure}[H]
    \centering
    \includegraphics[width=1\linewidth]{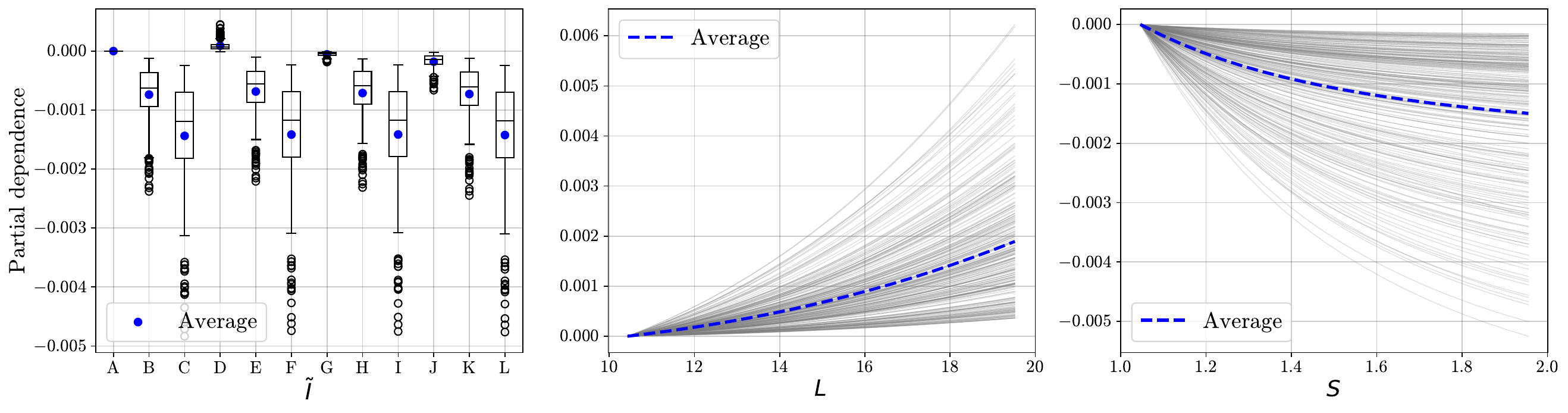}
    \caption{Partial dependence plots and individual conditional expectations for the cantilever beam bending problem.}
    \label{fig:cantilever_pdp}
\end{figure}

By looking at the basic strength of the material equation (see Eq.~(\ref{eq:cantilever}), it is clear that all variables interact with each other in how they affect the tip deflection. The PDP plots are shown in Fig.~\ref{fig:cantilever_pdp}. Note that the plot for the categorical variable is shown in the form of a boxplot due to its non-continuous nature (with cross-section A serving as the reference). This means that the partial dependence of each category on the model's predictions is represented by a boxplot, highlighting the distribution, central tendency, and variability of the model's predictions for each category. For both length and surface area, the trends are clear and align with engineering principles: increasing the length leads to greater tip deflection, whereas increasing the surface area results in reduced tip deflection. Moreover, the presence of interactions is evident since the ICE curves deviate from the centred value (\textit{i.e.}, partial dependence functions). For instance, the effect of increasing length on tip deflection depends on the type of cross-section and its area, as demonstrated by the variations in the ICE curves.

The PDP boxplot essentially shows the impact of the $\tilde{I}$, represented by various types of cross-sections. The result aligns with engineering principles, demonstrating that a more hollow cross-section (\textit{e.g.}, C is more hollow than B and A) leads to reduced deflection due to the increased moment of inertia. Additionally, when comparing different sections with similar levels of hollowness (\textit{e.g.}, C, F, I, and J), the impact on tip deflection is approximately the same. The relative change in impact is more significant when the level of hollowness varies because it greatly affects the moment of inertia. It is worth noting that this observation also aligns with the correlation matrix $R_{1}^{cat}$ discussed earlier.

\begin{figure}[H]
    \centering
    \includegraphics[width=1\linewidth]{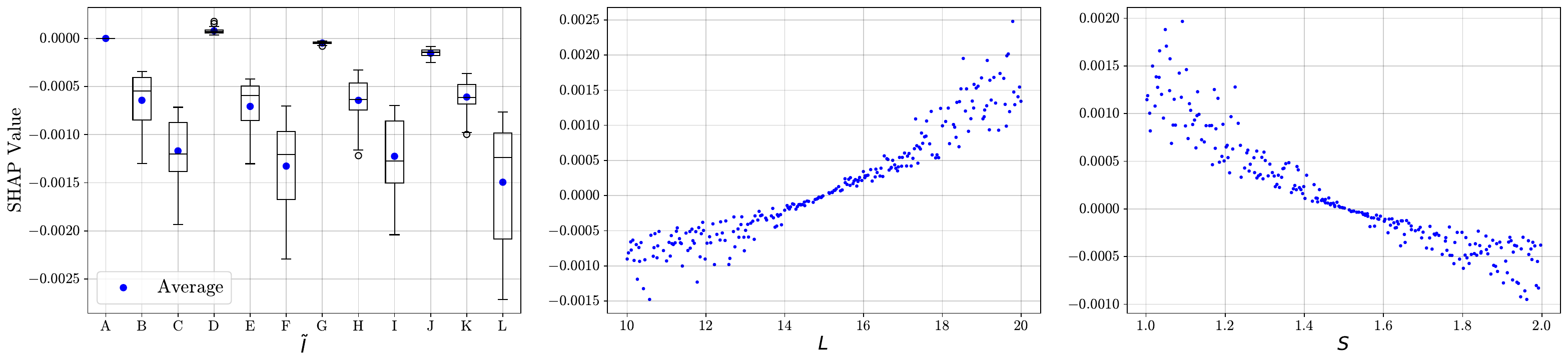}
    \caption{SHAP dependence plots for the cantilever beam bending problem.}
    \label{fig:cantilever_shap}
\end{figure}

The SHAP dependence plots offer different viewpoints but still reveal similar insights illustrated in Figure~\ref{fig:cantilever_shap}. 
It is worth noting that the sign of the SHAP value only corresponds to the deviation from the mean or the centre. Thus, it is the trend that should be interpreted. The scattered SHAP values when plotted versus $L$ and $S$ indicated again the interactions between the three variables in affecting the tip deflection. The grouping of the unique values of  $\tilde{I}$, representing the normalized moment of inertia, is also clearly observable from the SHAP dependence plots. 

\begin{figure}[H]
    \centering
    \subfigure[Based on PDP]{
    \includegraphics[width=0.45\linewidth]{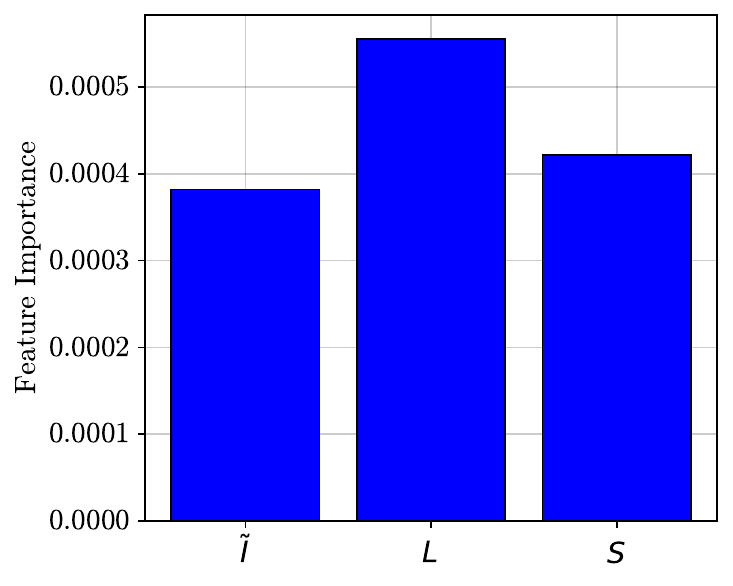}
    }
    \subfigure[Based on SHAP Value]{
    \includegraphics[width=0.45\linewidth]{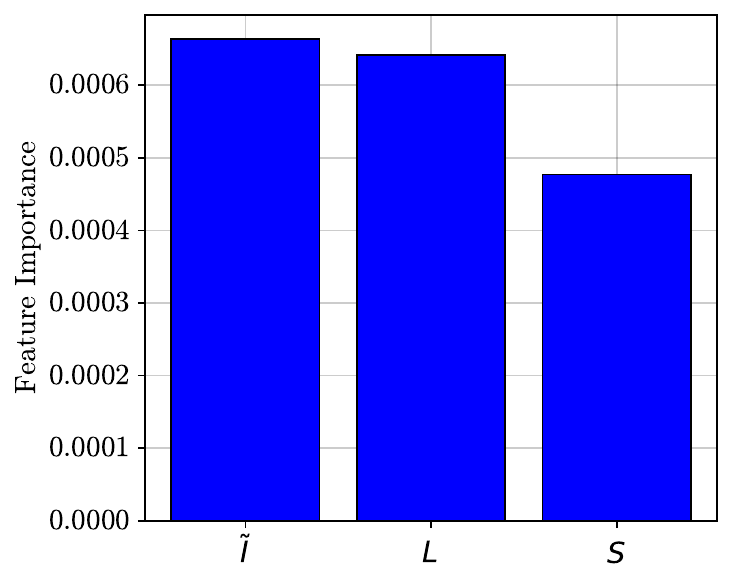}
    }
    \caption{Feature importance of the cantilever beam bending problem.}
    \label{fig:cantilever_feature_importance}
\end{figure}

Finally, the feature importance plots are shown in Fig.~\ref{fig:cantilever_feature_importance}. Both metrics agree that the length of the cantilever beam is a significant factor, given the range of continuous inputs and the type of categorical variable. However, SHAP suggests that $L$ and $\tilde{I}$ have roughly equal importance, whereas PDP identifies $L$ as the most important variable, with $\tilde{I}$ and $S$ following as the next most important variables, showing similar levels of importance. Like PDP, the confidence intervals of the $1$D models in Fig.~\ref{fig:cantilever_conf} seems to indicate similar importance for $\tilde{I}$ and $S$ and a more important $L$.
In other words, both importance metrics do not fully agree on the importance level of input variables. This difference is due to the distinct perspectives of the two metrics, as previously mentioned in the continuous 10-variable wing weight problem.
Although there is a difference in feature importance, it is important to emphasize that both SHAP and PDP/ICE accurately capture the actual trend of the problem, as previously discussed through the dependence plots.

This example illustrates that the explainability module in SMT-EX effectively complements the predictive model by providing a tool that helps users understand the impact of continuous and categorical variables.

\section{Conclusion}
SMT is a flexible Python-based surrogate modelling tool designed to handle a wide range of input problems, including those involving categorical variables. To further enhance its utility, this paper introduces an explainability module for the SMT tool, called SMT-EX. This module enables users to explore the constructed surrogate models, providing valuable insights into their behaviour and predictions. Once the model is built, the explainability module can be easily accessed, enhancing user interaction with the model. Currently, SMT-EX incorporates three explainability methods: SHAP, PDP, and ICE. This enables SMT users to gain deeper insights into the characteristics of the constructed surrogate model, facilitating a thorough understanding of the relationships between input variables and the output. \textcolor{black}{Besides, we have also implemented Sobol' indices as a ``conventional'' means for GSA and uncertainty quantification based on conformal prediction. The conformal prediction itself serves as a method to make the method more ``transparent'', similar to the way GP provides the uncertainty estimate. As such, the availability of uncertainty estimates also falls under the umbrella of explainability.} Leveraging the capabilities of SMT, our software effectively manages both mixed-categorical and continuous variables. We advocate for a paradigm shift in how engineers perceive their surrogate models—viewing them not merely as predictive tools but as insightful frameworks that reveal underlying relationships. This shift can significantly expand their understanding and prove beneficial for future applications. In this regard, we developed SMT-EX as a tool that can help scientists and engineers toward that end. 

We demonstrated the capabilities of SMT-EX on two test problems: a mixed-categorical beam problem and a 10-variable wing weight problem, where all variables are continuous. Initially, we constructed an accurate GPR model, as accuracy is fundamental for providing meaningful explainability. Subsequently, we activated the explainability module, which yielded insights through visualizations and sensitivity values generated by SHAP, PDP, and ICE methods. This validation is essential for confirming the utility of the knowledge produced, and we found that our results align with established engineering insights for two well-known problems in mechanics and aircraft design.


\section*{Acknowledgement}
The work of Pramudita Satria Palar and Mohammad Daffa Robani was supported by Institut Teknologi Bandung through the International Research Collaboration for New and Renewable Energy Program. The work of Saves Paul and Joseph Morlier is part of the activities of ONERA - ISAE - ENAC joint research group. 
We thank Rémi Lafage for its constant help and for being the administrator of the SMTorg open-source organization on GitHub.
The research presented in this paper has been performed in the framework of the COLOSSUS project (Collaborative System of Systems Exploration of Aviation Products, Services and Business Models) and has received funding from the European Union Horizon Europe program under grant agreement n${^\circ}$ 101097120. The authors acknowledge the research project MIMICO funded in France by the Agence Nationale de la Recherche (ANR, French National Research Agency), grant number ANR-24-CE23-0380. 
\bibliography{main}

\end{document}